\documentclass[10pt, a4paper]{article}
\usepackage{lrec}
\usepackage{multibib}
\newcites{languageresource}{Language Resources}
\usepackage{soul}

\usepackage{graphicx}
\usepackage{tabularx}
\usepackage{hyperref}

\usepackage{float}
\usepackage{amsmath}
\usepackage{amssymb}
\usepackage{multirow}

\usepackage{colortbl}
\usepackage{hhline}

\newcommand{\TC}[1]{\textcolor{black}{#1}}
\newcommand{\OC}[1]{\textcolor{black}{#1}}

\title{A Dataset Independent Set of Baselines\\ for Relation Prediction in Argument Mining}

\name{Oana Cocarascu\textsuperscript{1}, Elena Cabrio\textsuperscript{2}, Serena Villata\textsuperscript{2}, Francesca Toni\textsuperscript{1}}

\address{\textsuperscript{1}Imperial College London, UK \\
         \textsuperscript{2} Universit\'e C\^ote d'Azur, CNRS, Inria, I3S, France\\
         \{oana.cocarascu11, f.toni\}@imperial.ac.uk, \{elena.cabrio, serena.villata\}@unice.fr\\\\}

\abstract{
Argument Mining is the research area which aims at extracting argument components and predicting argumentative relations \OC{(i.e., \textit{support} and \textit{attack})} from text. 
\OC{In particular,} numerous approaches have been proposed in the \OC{literature} to predict the relations holding between the arguments, and application-specific annotated resources were built for this purpose. Despite the fact that these resources have been created to experiment on the same task, the definition of a single relation prediction method to be successfully applied to a significant portion of these datasets is an open research problem in Argument Mining. This means that none of the methods proposed in the literature can be easily ported from one \TC{resource} to another. In this paper, we address this problem by proposing a set of dataset independent strong \OC{neural} baselines which obtain homogeneous results on all the datasets proposed in the literature for the \OC{argumentative} relation prediction task. \OC{Thus,} our baselines can be employed by the Argument Mining community to compare more effectively how well a method performs on the \OC{argumentative} relation prediction task. \\ \newline }

\begin{document}

\maketitleabstract

\section{Introduction}
\textit{Argument(ation) Mining} (AM) is ``the general task of analyzing discourse on the pragmatics level and applying a certain argumentation theory to model and automatically analyze the data at hand''~\cite{Habernal:2017:AMU:3097255.3097259}. Two tasks are crucial~\cite{DBLP:journals/ijcini/PeldszusS13,DBLP:journals/toit/LippiT16,DBLP:conf/ijcai/CabrioV18}: \textit{1)} argument component detection within the input natural language text aiming at the identification of arguments (claim, premises, and their textual boundaries); and \textit{2)} relation prediction aiming at the prediction of the relations between the argumentative components identified in the first stage (support, attack).

Despite the \OC{high volume} of approaches tackling the relation prediction task with satisfying results (see~\cite{DBLP:conf/ijcai/CabrioV18} for the complete list), a problem arises: these solutions heavily rely on the peculiar features of the dataset taken into account for the experimental setting and are hardly portable from one application domain to another. 

On the one side, this issue can be explained by the huge number of heterogeneous application domains where argumentative text may be analysed (e.g., online reviews, blogs, political debates, legal cases). On the other side, it represents a drawback for the comparison of the different approaches proposed in the literature, which are often presented as solutions addressing the relation prediction task from a dataset independent point of view. A side drawback for the AM community is therefore a lack of big annotated resources for this task, as most of them cannot be successfully reused. 
In this paper, we tackle this issue by proposing a set of strong cross-dataset baselines based on different neural architectures. Our baselines are shown to perform homogeneously over all the datasets proposed in the literature for the relation prediction task in AM, \OC{differently from} what is achieved by the single methods proposed in the literature. The contribution of our proposal is to bestow the AM community with a set of strong cross-dataset baselines to compare with in order to demonstrate how well a relation prediction method \OC{for AM} performs.

The majority of the datasets containing argumentative relations target only two types of relations: attack and support. We define neural models to address the binary classification problem, analysing, \TC{to the best of our knowledge}, all available datasets for this task ranging from persuasive essays to user-generated content, to political speeches. Given two arguments, we are interested in determining the relation between the first, called \emph{child} argument, and the second, called \emph{parent} argument, by means of a neural network. \OC{For example, the child argument \emph{People know video game violence is fake} attacks the parent argument \emph{Youth playing violent games exhibit more aggression}.} Each of the two arguments is represented using embeddings as well as other features.  

\TC{Current papers that target \OC{AM} propose different neural networks for different datasets. In this paper, we propose several neural network architectures and perform a systematic evaluation of these architectures on different datasets for the relation prediction in argument mining. We provide a broad comparison of different deep learning methods for a large \OC{number} of datasets for the relation prediction \OC{in AM}, an important and still widely open problem.} \OC{Concretely,} we propose \TC{four} neural network architectures for the classification task, \TC{two} concerned with the way child and parent are passed through the network (\emph{concat} model \TC{and \emph{mix} model}), an autoencoder, and an attention-based model.

In the remainder of the paper, Section~\ref{am-datasets} presents the datasets used in the experiments, along with their main linguistic features. Section~\ref{am-dl} describes the \TC{features, and the} deep learning models. \TC{We report the performance of the proposed models in} Section~\ref{am-experiments} Conclusions for the paper are in Section \ref{concl}

\section{Relation-based AM datasets} \label{am-datasets}

In this section, we describe the datasets that we used to compute our baselines\footnote{We do not consider the two legal datasets built for relation prediction by~\cite{Mochales2011} and~\cite{Teruel:18} because the former is not available and the latter has a low inter-annotator agreement.}. Datasets statistics can be found in Table~\ref{data-stats}\footnote{For more details about the single datasets, we refer the reader to the related publication.}. \TC{We focused on these datasets as they were specially created for the relation prediction in AM or they can be easily transformed to be used for this task.}

\begin{table}
\centering
\begin{tabular}{|l|c|c|c|}
\hline \textbf{Dataset} & \textbf{ID} & \textbf{\# attacks} & \textbf{\# supports} \\ \hline
\textbf{Essays} & essay & 497 & 4841 \\ \hline
\textbf{Microtexts} & micro&  108 & 263 \\ \hline
\textbf{Nixon-Kennedy} & nk & 378 & 353 \\ \hline
\textbf{Debatepedia} & db & 141 & 179 \\ \hline
\textbf{IBM} & ibm & 1069 & 1325 \\ \hline
\textbf{ComArg} & com & 296 & 462 \\ \hline
\textbf{Web-content} & web & 1301 & 1329 \\ \hline
\textbf{CDCP} & cdcp & 0 & 1220 \\ \hline
\textbf{UKP} & ukp & 5935 & 4759  \\
 \hline
 \textbf{AIFdb} & aif & 9854 & 7543  \\
 \hline
\end{tabular}
\caption{Summary of datasets.}
\label{data-stats}
\end{table}

\begin{itemize}
\item \textit{Persuasive essays} \cite{Stab:17}: a corpus of 402 persuasive essays annotated with discourse-level argumentation structures. \TC{The major claim represents the author's standpoint on the topic, which is supported or attacked by claims which in turn can be supported or attacked by premises.} An example of (a part of) an essay is below:

\textit{Ever since researchers at the Roslin Institute in Edinburgh cloned an adult sheep, there has been an ongoing debate about whether cloning technology is morally and ethically right or not. Some people argue for and others against and there is still no agreement whether cloning technology should be permitted. However, as far as I’m concerned, $[$cloning is an important technology for humankind$]_{\text{MajorClaim1}}$ since $[$it would be very useful for developing novel cures$]_\text{Claim1}$.} 
\textit{First, $[$cloning will be beneficial for many people who are in need of organ transplants$]_\text{Claim2}$. $[$Cloned organs will match perfectly to the blood group and tissue of patients$]_\text{Premise1}$
since $[$they can be raised from cloned stem cells of the patient$]_\text{Premise2}$.}

In this example, both Claim1 and Claim2 support the Major Claim, Premise1 supports Claim2 and Premise2 supports Premise1.

\item \textit{Microtexts} \cite{Peldszus:15}: a corpus of 112 microtexts covering controversial issues. 
\OC{We focus on normal supports and rebut attacks only. The dataset has in addition examples and rebut attacks but we discard the former due to them being rarely used and the latter because we are not interested in attacks to inferences.} An example of a microtext can be seen in Figure \ref{microtext}. Here, the second segment rebuts the first segment and the third segment undercuts the link between the second segment and the first segment. Segments four and five jointly support the main claim.

\begin{figure}[H]
    \centering
    \includegraphics[width=0.5\textwidth]{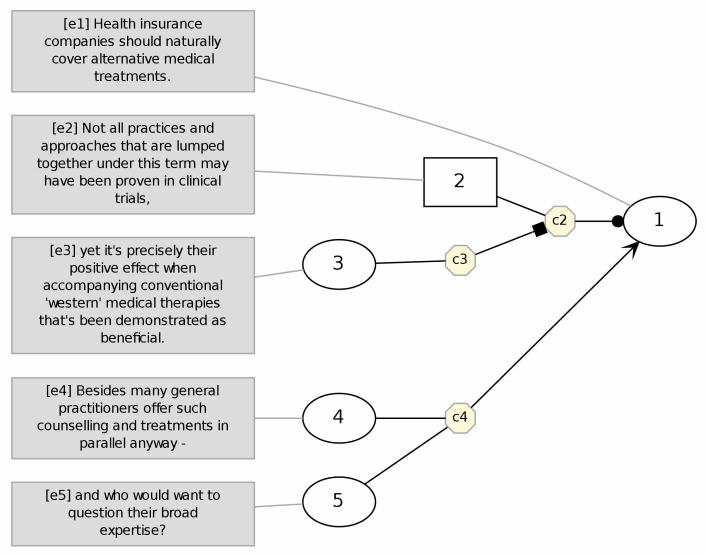}
	\caption{An example of microtext and the associated argumentation graph.
    \label{microtext}}
\end{figure}

\item \textit{Nixon-Kennedy debate} \cite{Menini:18}: a corpus from the Nixon-Kennedy presidential campaign covering five topics: Cuba, disarmament, healthcare, minimum wage and unemployment. Below are two examples from the dataset:

\emph{We could have tried to inject ourselves into the Congo without honoring our commitments to the United Nations charter, just as Khrushchev seems to be trying to do. We could have turned Cuba into a second Hungary. But we can be eternally grateful that we have a man in the White House who did none of these things}  \textbf{supports} \emph{I don't take the views that the only alternative to a dictator is a Communist dictator. If the United States had just had its influence, and at that time the United States was extremely powerful in Cuba, it seems to me we could have persuaded Mr. Batista to hold free elections at the time he was permitted to go and permit the Cuban people to make their choice instead of letting Castro seize power through revolution. I think we are going to have a good deal of trouble in the future with Castro through all of Latin America.}

\emph{They are afraid of diplomatic policies that teeter on the brink of war. They are dismayed that our negotiators have no solid plans for disarmament. And they are discouraged by a philosophy that puts its faith in swapping threats and insults with the Russians} \textbf{attacks} \emph{It's a typically specious and frivolous maneuver. We have made a good-faith effort to advance the - advance toward disarmament - and make some progress by having a meeting of the Disarmament Commission. Now, when they make a proposal like this, it's a cynical attempt to prevent progress, that's what it is it shows that they don't really want disarmament.}

\item \textit{Debatepedia} \cite{Cabrio:14}:  a corpus from the two debate platforms Debatepedia\footnote{\url{http://idebate.org/debatabase}} and ProCon\footnote{\url{http://www.procon.org/}}. Below are two examples from the dataset:

\emph{Research studies have yielded the conclusion that the effect of violent media consumption on aggressive behavior is in the same ballpark statistically as the effect of smoking on lung cancer, the effect of lead exposure on children's intellectual development and the effect of asbestos on laryngeal cancer.} \textbf{supports } \emph{Violent video games are real danger to young minds.} 

\emph{People know video game violence is fake.} \textbf{attacks}  \emph{Youth playing violent games exhibit more aggression.}

\item \textit{IBM} \cite{Slonim:17}:  a dataset from topics randomly selected from the debate motions database at the International Debate Education Association (IDEA)\footnote{\url{ http://idebate.org/}}. Below are two examples from the dataset:

\emph{Children with many siblings receive fewer resources.} \textbf{supports} \emph{This house supports the one-child policy of the republic of China.}

\emph{Virtually all developed countries today successfully promoted their national industries through protectionism} \textbf{attacks} \emph{This house would unleash the free market.}

\item \textit{ComArg} \cite{Boltuvzic:14}: a corpus of online user comments from ProCon and IDEA. We combine the two types of attacks (explicit and vague/implicit attacks) and the two types of supports (explicit and vague/implicit arguments). Below are two examples from the dataset:

\emph{Religion should stay out of the public square, except when people exercise their right to the freedom of speech an expression. Having Under God in the pledge forces all people to pledge allegiance to a higher power they may not believe in. The separation of Church and State should disallow such favoritism. Can anyone fathom the reaction of believers if it said: One Nation, created by a big bang and inhabited by evolved creatures.... ?} \textbf{supports} \emph{Removing under god would promote religious tolerance} 

\emph{Atheism doesn't mean the absence of religion - it means the absence of a god in one's belief system.  Certain forms of Buddhism, for example are atheistic.  Therefore, requiring a statement of belief in a god is unconstitutionally preferring a majority religious belief over a minority one.  The point of the Pledge is to state allegiance to the flag and country.  If one believes in a god, there are many, many other forums in which to express that belief without imposing it on others.} \textbf{attacks} \emph{America is based on democracy and the pledge should reflect the belief of the American majority.}

\item \textit{Web-content dataset} \cite{Carstens:15}: a dataset of arguments adapted from the Argument Corpus~\cite{Walker:12}, plus arguments from news articles, movies, ethics and politics. Below are two examples from the dataset:

\emph{i agree did not like this either in fact i stopped watching once waltz was killed because i just didnt care anymore} \textbf{supports} \emph{after all the attention and awards etc and an imdb rating of i was so shocked to finally see this film and have it be so bad}

\item \emph{samsung note it has a bigger screen and a somewhat faster processor} \textbf{attacks} \emph{htc one it is currently the best one in the market good quality superb specs}

\item \textit{CDCP} \cite{Park:18}:  a dataset consisting of support arguments only from user comments regarding Consumer Debt Collection Practices from an eRulemaking website\footnote{\url{http://regulationroom.org}}. Below are two examples:

\emph{sundays really are when most people are spending whatever little time they have left before the workweek with friends and family} \textbf{supports} \emph{i do not conduct business on sundays}

\item \emph{a robo-call that tells you that you have a message or an account update, and the only way to get it is to call a special number with an extension, but when you call, it is just the same message asking where your payment is, is a waste of the consumer's time and the consumer's cellular resources (two phone calls, one received, one sent} \textbf{supports} \emph{i support these restrictions on robo-calling and any calls during the work hours}

\item \textit{UKP} \cite{Stab:18}: a dataset of arguments from online comments on \TC{8} controversial issues: \TC{abortion, cloning, death penalty, gun control, minimum wage, nuclear energy, school uniforms, marijuana legalization}. In this dataset, one of the arguments is represented by the topic. Below are two examples:

\emph{Dr. Strouse has seen both the benefits and risks of cannabis use and is well-versed in the emerging scientific evidence regarding the effectiveness of cannabinoids in a variety of medical conditions and pain states, as well as epidemiologic evidence of legalized marijuana's connection to a reduction in prescription drug use and opioid-related deaths} \textbf{supports} \emph{marijuana legalization} 

\emph{Would you want to live in a neighborhood filled with people who regularly smoke marijuana} \textbf{attacks} \emph{marijuana legalization}

For our experiments, we modify the parent text from \emph{topic} to a default seen as the natural language template \emph{topic is good}. Hence from the previous example, we would have an argument for and an argument against ``marijuana legalization is good".

\item \textit{AIFdb} \cite{Bex:13,Chesnevar:06,Rahwan:09,Reed:08a}: \TC{a corpus of argument maps which follows the structure defined by AIF \cite{Lawrence:12}. We select the following datasets from AIFdb and keep the English texts only: AraucariaDB, DbyD Argument Study, Expert Opinion and Positive Consequences, Internet Argument Corpus, Mediation (here we compiled the following datasets: Dispute mediation, Dispute mediation: excerpts taken from publications, Mock mediation, Therapeutic, Bargaining, Meta-talk in mediation), Opposition (here we compiled the following datasets: Language Of Opposition Corpus 1, Android corpus, Ban corpus, Ipad corpus, Layoffs corpus, Twitter corpus). We map the original set of relations to 2 classes as follows: CA-nodes are mapped to attack and RA- and TA-nodes are mapped to support.} Below are two examples form the dataset: \\
\emph{the water temperature is perfect} \textbf{supports} \emph{Burleigh Heads Beach is the best.} \\
\emph{We should implement Zoho, because it is cheaper than MS Office} \textbf{attacks} \emph{We should implement OpenOffice.}

\end{itemize}

\TC{In terms of results reported on the datasets we have conducted our experiments on, most works perform a  cross-validation evaluation or, in the case of datasets consisting of several topics, the models proposed are trained on some of the topics and tested on the remaining topics.} 

\TC{For the \emph{essay} dataset, an Integer Linear Programming model was used to achieve 0.947 $F_1$ for the support class and 0.413 $F_1$ for the attack class on the testing dataset using cross-validation to select the model \cite{Stab:17}. Using SVM, 0.946 $F_1$ for the support class and 0.456 $F_1$ for the attack class were obtained \cite{Stab:17}. Using a modification of the Integer Linear Programming model to accommodate the lack of some features used for the \emph{essay} dataset but not present in the \emph{micro} dataset, 0.855 $F_1$ was obtained for the support class and 0.628 $F_1$ for the attack class.} \TC{On the \emph{micro} dataset, an evidence graph model was used to achieve 0.71 $F_1$ using cross-validation \cite{Peldszus:15}.} \TC{On the \emph{nk} dataset, 0.77 $F_1$ for the attack class and 0.75 $F_1$ for the support class were obtained using SVM and cross-validation \cite{Menini:18}.} \TC{SVM accuracy results on the testing dataset using coverage (i.e. number of claims identified over the number of total claims) were reported in \cite{Slonim:17} as follows: 0.849 accuracy for 10\% coverage, 0.740 accuracy for 60\% coverage, 0.632 accuracy for 100\% coverage.} \TC{Random Forests were evaluated on the \emph{web} and \emph{aif} datasets using cross-validation, achieving 0.717 $F_1$ and 0.831 $F_1$, respectively \cite{Carstens:17}.} \TC{Structured SVMs were evaluated in a cross-validation setting on the \emph{cdcp} and \emph{ukp} datasets using various types of factor graphs, full and strict \cite{Niculae:17}. On the \emph{cdcp} dataset, $F_1$ was 0.493 on the full graph and 0.50 on the strict graph whereas on the \emph{ukp} dataset, $F_1$ was 0.689 on the full graph and 0.671 on the strict graph.} \TC{No results on the two-class datasets were reported for \emph{db}, \emph{com}, and \emph{ukp} datasets. The results on \emph{ukp} treat either supporting and attacking arguments as a single category or considering three types of relations: support, attack, neither. The latter type of reporting results on three classes is also given on the \emph{com} dataset.}

\section{Neural baselines for relation prediction} \label{am-dl}

In this section we describe the features used and the proposed neural models.

\subsection{Features} \label{features}

We use four types of features: embeddings, textual entailment, sentiment features, and syntactic features, computed for \emph{child} and \emph{parent}, respectively. We refer to the last three types of features as \emph{standard} features.

Word embeddings are distributed representations of texts in an n-dimensional space. We add a feature of entailment \OC{from child to parent} representing the class (entailment, contradiction, or neutral) obtained using AllenNLP\footnote{\url{https://allennlp.org/models}}, a textual entailment model based on a decomposable attention model \cite{Parikh:16}.
The features related to sentiment are based on manipulation of SentiWordNet \cite{Esuli:06} and the sentiment of the entire text analysed \OC{using the VADER sentiment analyser \cite{Hutto:14}}. Every WordNet synset \cite{Miller:95} \OC{can be} associated to three scores describing how objective, positive, and negative it is. For every word in the text \OC{(child and parent, respectively}), we select its first synset and compute its positive score and its negative score. \OC{In summary,} the features related to sentiment for a text \emph{t} that consists of \OC{\emph{n} words, i=1..n,} are the following: (i) sentiment score ($\sum_{w_i} pos\_score(w_i) - neg\_score(w_i)$), (ii) number of positive/negative/neutral words in \emph{t} (a word is neutral if \emph{not}($pos\_score(w_i) > 0$ and $neg\_score(w_i) > 0$)), (iii) sentiment polarity class and score of \emph{t}.
Syntactic features consist of text statistics and word statistics with respect to the POS tag: number of words, nouns, verbs, first person singular, second person singular and plural, third person singular and plural, first person plural, modals, modifiers (number of adverbs plus the number of adjectives), and lexical diversity (number of unique words divided by the total number of words in \OC{text} \emph{t}).

\subsection{Neural Architectures} \label{sec:neural-arch}

We describe the \TC{four} \OC{neural} architectures we propose for determining the argumentative relation (attack or support) holding between two texts \OC{(see Figures 1-4)}. \OC{For all,} the number of the hidden layers and their sizes are the ones that performed the best. We report only \OC{on configurations of the architectures as given in Section \ref{sec:neural-arch} as these were the best performing.} However, we experimented with 1 and 2 hidden layers, and hidden layer sizes of 32, 64, 128, and 256, \TC{trying all possible combinations in order to obtain the best configurations, and limiting to 2 hidden layers due to the small size of the data}. For our models, we use GRUs~\cite{Cho:14}. Various works have compared LSTMs and GRUs but~\cite{Chung:14,Jozefowicz:15} did not obtain conclusive results as to which type is better, suggesting that the design choice is dependant on the dataset and task. We focus on GRUs as they take less time to train and are more efficient as LSTMs have more parameters.

\subsubsection{Concat model}

\begin{figure}
    \centering
    \includegraphics[width=0.47\textwidth]{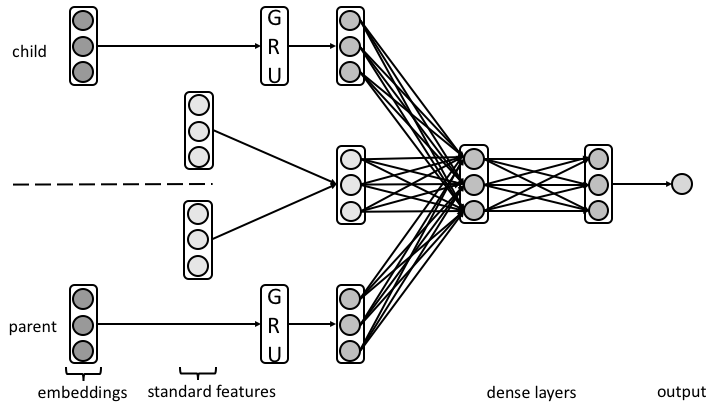}
	\caption{The concat architecture.
    \label{fig:concat_dl}}
\end{figure}

In the concat model, each of the child and parent embeddings is passed through a GRU. We concatenate the standard features of the child and parent. The merged standard vector is then concatenated with the outputs of the GRUs. The resulting vector is passed through 2 dense layers (of 256 neurons and 64 neurons, with sigmoid as activation function), and then to softmax to determine the argumentative relation. The concat model can be seen in Figure \ref{fig:concat_dl}.

\subsubsection{Mix model (M)}

\begin{figure}
    \centering
    \includegraphics[width=0.47\textwidth]{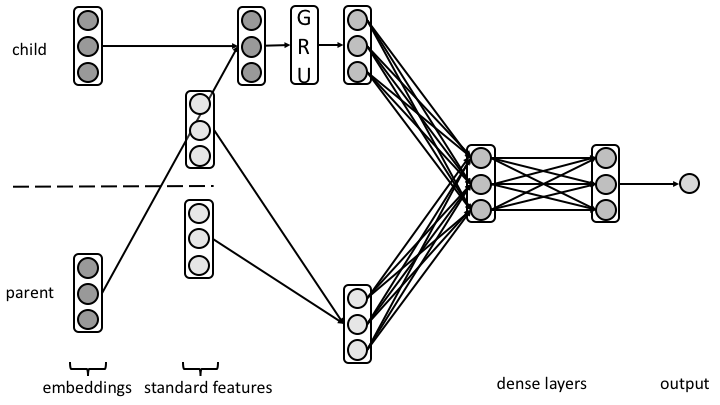}
	\caption{The mix architecture.
    \label{fig:mix_dl}}
\end{figure}

In the mix model, we first concatenate the child and parent embeddings and then pass them through a GRU, differently from the concatenation model where we pass each embedding vector through a GRU first. We concatenate the standard features that we obtain for the child and for the parent, respectively. The merged standard vector is then concatenated with the output of the GRU. From this stage, the network resembles the concatenation model: the resulting vector is passed through 2 dense layers (of 256 neurons and 64 neurons, with sigmoid as activation function), to be then finally passed to softmax to determine the argumentative relation. The mix model can be seen in Figure \ref{fig:mix_dl}.

\subsubsection{Autoencoder model}

\begin{figure}
    \centering
    \includegraphics[width=0.47\textwidth]{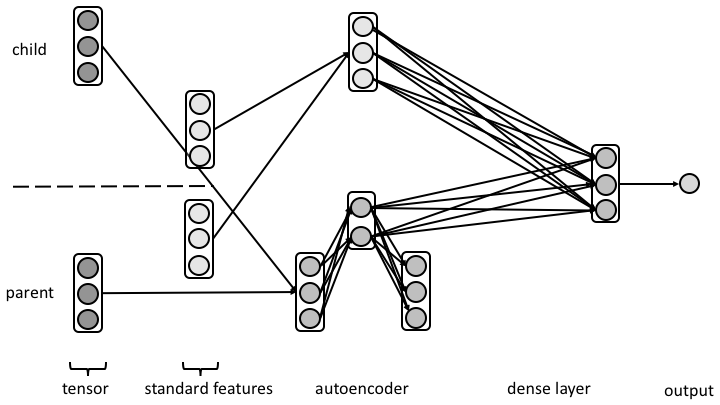}
	\caption{The autoencoder architecture.
    \label{fig:autoencoder_dl}}
\end{figure}

Autoencoders \cite{Hinton:06,Erhan:10} are unsupervised learning neural models which take a set of features as input and aim, through training, to reconstruct the inputs. Autoencoders can be used as feature selection methods to determine which features are redundant \cite{Wang:17,Han:17}.
We first concatenate the child and parent tensors, to obtain a vector of size $X$. We use an autoencoder with one hidden layer defined as: (i) an encoder function $f(X)=\sigma(XW^{(1)})$, and (ii) a decoder function $\sigma(f(X)W^{(2)})$, where  $W^{(1)}, W^{(2)}$ are the weight parameters in the encoder and decoder, respectively. The size of the hidden layer is 128. We use sigmoid as activation function in the autoencoder and binary cross entropy as loss function. We concatenate the standard features of the child and of the parent. The merged standard vector is then concatenated with the hidden layer in the autoencoder (Figure \ref{fig:autoencoder_dl}) which represents the encoded dataset as dimensionally reduced features. The resulting vector is passed through a single dense layers (of 32 neurons, with sigmoid as activation function), that is then passed to softmax.

\subsubsection{Attention model}

\begin{figure}
    \centering
    \includegraphics[width=0.47\textwidth]{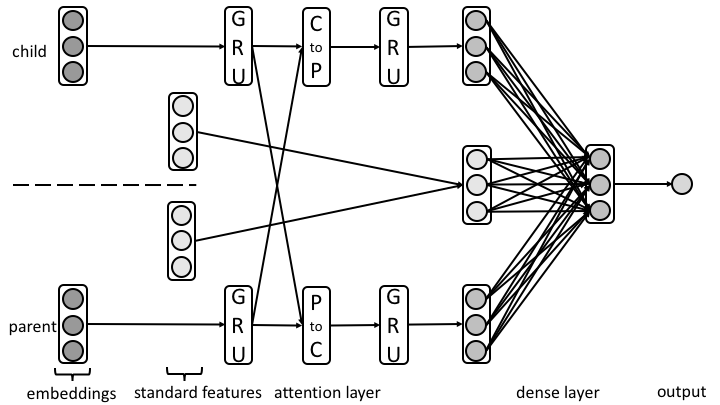}
	\caption{The attention architecture.
    \label{fig:attention_dl}}
\end{figure}

Inspired by the demonstrated effectiveness of attention-based models \cite{Yang:16,Vaswani:17}, we combine the GRU-based model with attention mechanisms.

Each of the child and parent embeddings is passed through a GRU. Let $C \in \mathbb{R}^{L_c \times d}$ be the output the GRU produces when reading $L_c$ words of Child $C$, and let $P \in \mathbb{R}^{L_p \times d}$ be the output the GRU produces when reading $L_p$ words of Parent $P$, \OC{where $d$ is the output dimension}. We compute attention in two directions: from Child $C$ to Parent $P$ and from $P$ to $C$. We illustrate the attention in one direction only. Let $s_{ij}$ be the similarity matrix between the $i$-th child word and the $j$-th parent word, $\alpha$ the attention weight, $c'_i$ the attended child vector, and $c''_i$ parent-aware representation of each child as follows:
\begin{flalign*} 
s_{ij}=W^{1 \times 2d} [c_{i:} ; p_{j:}]^{2d \times 1} + b^{1 \times 1}                            	&& 	s \in \mathbb{R}^{L_c \times L_p} \\
\alpha_{i:} = \text{softmax}(s_{i:})  		&&  	\alpha \in \mathbb{R}^{L_c \times L_p} \\
c'_{i:} = \alpha_{i:} p_{i:} 		&&	 c' \in \mathbb{R}^{L_c \times d} \\
c''_{i:} = [c_{i:} ; c'_{i:}] 	&&	 c'' \in \mathbb{R}^{L_c \times 2d}
\end{flalign*}  
where $W$ is a trainable weight vector and $[;]$ is vector concatenation across row. The weights vectors $W$ for the two directions are different. We concatenate the standard features of the child and of the parent. The merged standard vector is then concatenated with the outputs of the GRUs whose inputs are $c''$ and $p''$. The resulting vector is passed through a single dense layers (128 neurons, activation function: sigmoid), that is then passed to softmax (Figure \ref{fig:attention_dl}).

\begin{table*}
\centering
\arrayrulecolor{black}
\begin{tabular}{|l|l|l|c|c|c|c|c|c|c|c|c|c|c|c|}
\hline & & & essay & micro & db & ibm & com & web & cdcp & ukp & nk & aif & Avg & Mcr Avg\\ 
\hline

\multirow{16}{*}{\rotatebox[origin=c]{90}{non-neural baselines}} & \multirow{2}{*}{RF}  & $F_1$ A & & 0.24&0.22&0.25&0.03&0.27&-&0.22&0.43&0.31&0.246 & \multirow{2}{*}{0.467}  \\ 
& & $F_1$ S &   & 0.80&0.71&0.67&0.75&0.63&0.94&0.60&0.53&0.56&0.688 & \\ 
\cline{2-15}

& \multirow{2}{*}{RF}  & $F_1$ A & 0.32 & 0.24&0.40 & & 0.33&0.38&-&0.39&0.55&0.44&0.381 & \multirow{2}{*}{0.508} \\ 
& & $F_1$ S & 0.57 & 0.74&0.64 & & 0.67&0.59&0.85&0.53&0.59&0.54&0.636 & \\ 
\cline{2-15}

& \multirow{2}{*}{RF}  & $F_1$ A & 0.57 & 0.40&0.45&0.53&0.43 & & -&0.60&0.52&0.57&0.509 & \multirow{2}{*}{0.490} \\ 
& & $F_1$ S & 0.44 &  0.47&0.52&0.41&0.57 & & 0.51&0.45&0.50&0.38&0.472 & \\ 
\cline{2-15}

& \multirow{2}{*}{RF}  & $F_1$ A & 0.67 & 0.45&0.60&0.62&0.56&0.62&-&0.71&0.68 & &0.614& \multirow{2}{*}{0.335} \\ 
& & $F_1$ S & 0.01 & 0.02&0.17&0.01&0.04&0.24&0.01&0.00&0.00 & & 0.056 &   \\ 
\cline{2-15}

& \multirow{2}{*}{SVM}  & $F_1$ A & &  0.34&0.36&0.33&0.29&0.38&-&0.42&0.42&0.40&0.368 &   \multirow{2}{*}{0.503}  \\ 
& & $F_1$ S & &  0.71&0.67&0.65&0.67&0.59&0.84&0.57&0.56&0.49&0.639 &    \\ 
\cline{2-15}

& \multirow{2}{*}{SVM}  & $F_1$ A & 0.52 & 0.42&0.37 & & 0.43&0.50&-&0.59&0.33&0.49&0.456 & \multirow{2}{*}{0.473} \\ 
& & $F_1$ S & 0.48 & 0.51&0.50 & & 0.45&0.47&0.61&0.43&0.53&0.42&0.489 & \\ 
\cline{2-15}

& \multirow{2}{*}{SVM}  & $F_1$ A & 0.49 & 0.35 & 0.39 & 0.39 & 0.38 & & - & 0.56 & 0.57 & 0.520 & 0.456 & \multirow{2}{*}{0.498
} \\ 
& & $F_1$ S & 0.50 & 0.54 & 0.52 & 0.59 & 0.60 & & 0.67 & 0.46 & 0.47 & 0.500 & 0.539 & \\ 
\cline{2-15}

& \multirow{2}{*}{SVM}  & $F_1$ A & 0.61 & 0.40 & 0.60 & 0.46 & 0.57 & 0.61 & - & 0.64 & 0.68 & & 0.571 & \multirow{2}{*}{0.431} \\ 
& & $F_1$ S & 0.35 & 0.50 & 0.22 & 0.57 & 0.04 & 0.24 & 0.40 & 0.30 & 0.00 &  & 0.291 & \\ 
\hline
\end{tabular}
\arrayrulecolor{black}
\caption{Results on the datasets with attack (A) and support (S) relations. $F_1$ A stands for the $F_1$ measure of the attack relation and $F_1$ S stands for the $F_1$ measure of the support (S) relation. RF stands for Random Forests. \OC{The blanks represent the training dataset. The Average (Avg) \OC{and the Macro (\OC{Mcr}) Avg do} not include the results of the dataset used for training.}}
\label{non-neural-baselines}
\end{table*}
\section{Experimental results} \label{am-experiments}

\subsection{Non-neural baselines}

We have used for training the larger datasets, \emph{aif}, \emph{essay}, \emph{ibm} and \emph{web}. We resampled the minority class from the \emph{essay} dataset and used our models on the oversampled dataset. We did not used for training the \emph{ukp} dataset as the parent is a topic instead of an  argument. The models are then tested on the remaining datasets with the average being computed on testing datasets. We report the $F_1$ performance of the \emph{attack} class (A) and the \emph{support} class (S).
\OC{Table \ref{non-neural-baselines} shows the results for the non-neural baselines.} We used Random Forests (RF) \cite{Breiman:01} with 15 trees in the forest and gini impurity criterion and \TC{SVM with linear kernel using LIBSVM \cite{Chang:11}, obtained as a result of performing a grid search} as it is the most commonly used algorithm in the works that experiment on the datasets we considered \cite{Slonim:17,Boltuvzic:14,Carstens:17,Menini:18,Niculae:17}. On top of the \emph{standard} features used for our neural models, \OC{for the baselines} we added the following features: TF-IDF, number of common nouns, verbs and adjectives between the two texts as in \cite{Menini:18},  a different sentiment score $\frac{\text{nr\_pos}-\text{nr\_neg}}{\text{nr\_pos}+\text{nr\_neg}+1}$ as in \cite{Slonim:17}, all features being normalized.

\subsection{Neural baselines with non-contextualised word embeddings} \label{non-context}

Table~\ref{syntactic-100} shows the best baselines for relation prediction in AM.  
For our models, we experimented with two types of embeddings: GloVE \cite{Pennington:14} (300-dimensional) and FastText (FT) \cite{Joulin:16,Mikolov:18} (300-dimensional). We used pre-trained word representations in all our models. \TC{We used 100 as the sequence size as we noticed that there are few instances with more than 100 words}. We used a batch size of 32 and trained for 10 epochs (as a higher number of epochs led to overfitting). 
We report the results using embeddings and \emph{syntactic} features and the results with \emph{all} the features presented in Section \ref{features} We also conducted a feature ablation experiment (with embeddings being always used) and observed that syntactic features contribute the most to performance, with the other types of features bringing small improvements when used together only with embeddings. 
\TC{In addition, we have run experiments using two datasets for training to test whether combining two datasets improves performance. During training, we used one of the large datasets (\emph{aif}, \emph{essay}, \emph{ibm}, \emph{web}) and one of the remaining datasets (represented as blanks in the Table\footnote{For readability, blanks represent the training datasets.}).}

\begin{table*}
\centering
\arrayrulecolor{black}
\begin{tabular}{|l|l|l|c|c|c|c|c|c|c|c|c|c|c|c|}
\hline & & & essay & micro & db & ibm & com & web & cdcp & ukp & nk & aif & Avg & Mcr Avg\\ 
\hline
\multirow{10}{*}{\rotatebox[origin=c]{90}{embeddings + syntactic}} & \multirow{2}{*}{M FT} & $F_1$ A & 0.52 & 0.40 & 0.58 & 0.50 & 0.52 &  & -    & 0.58 & 0.52 &  &  0.517 & \multirow{2}{*}{0.521}  \\ 
& & $F_1$ S & 0.47 & 0.50 & 0.61 & 0.54  & 0.54 &  & 0.60 & 0.42 & 0.53 &  &  0.526 &  \\
\cline{2-15}
& \multirow{2}{*}{C FT} & $F_1$ A &  & 0.36 &  & 0.45 & 0.47 & 0.52 & -    & 0.65 & 0.46 & 0.52 &  0.490 & \multirow{2}{*}{0.52} \\ 
& & $F_1$ S &  & 0.64 &  & 0.55 & 0.67 & 0.53 & 0.72 & 0.34 & 0.48 & 0.47 &  0.550 &  \\ 
\cline{2-15}
& \multirow{2}{*}{C G}  & $F_1$ A &  & 0.35 & 0.43 & 0.48 & 0.31 & 0.45 & -    & 0.58 &  & 0.43 &  0.433 & \multirow{2}{*}{0.526}  \\ 
& & $F_1$ S &  & 0.71 & 0.68 & 0.58 & 0.70 & 0.54 & 0.77 & 0.47 &  & 0.50 &  0.619 &  \\ 
\cline{2-15}
& \multirow{2}{*}{A G}  & $F_1$ A &  & 0.37 & 0.58 &  & 0.53 & 0.53 & -    & 0.61 & 0.59 & 0.55 &  0.537 & \multirow{2}{*}{0.526} \\ 
& & $F_1$ S &  & 0.61 & 0.60 &  & 0.42 & 0.50 & 0.72 & 0.43 & 0.38 & 0.47 &  0.516 &  \\ 
\cline{2-15}
& \multirow{2}{*}{A G}  & $F_1$ A &  & 0.36 & 0.48 & 0.43 & 0.39 &  & -    & 0.52 & 0.45 & 0.51 & 0.449 & \multirow{2}{*}{\textbf{0.544}}  \\ 
& & $F_1$ S &  & 0.75 & 0.66 & 0.62 & 0.68 &  & 0.79 & 0.52 & 0.56 & 0.54 &  0.640 &  \\ 
\hline

\multirow{18}{*}{\rotatebox[origin=c]{90}{all features}} & \multirow{2}{*}{M G}  & F1 0 & 0.49 & 0.41 & 0.51 &  & 0.50 & 0.52 & - & 0.57 & 0.49 &  &  0.499 & \multirow{2}{*}{0.517}  \\ 
& & F1 1 & 0.49 & 0.63 & 0.60 &  & 0.52 & 0.45 & 0.69 & 0.37 & 0.54 &  &  0.536 &  \\ 
\cline{2-15}
& \multirow{2}{*}{C G}  & F1 0 & 0.42 & 0.33 & 0.52 & 0.35 & 0.48 &  & - & 0.49 & 0.43 &  &  0.431 &\multirow{2}{*}{0.515}  \\ 
& & F1 1 & 0.54 & 0.68 & 0.63 & 0.66 & 0.53 &  & 0.75 & 0.46 & 0.55 &  &  0.600 &  \\ 
\cline{2-15}
& \multirow{2}{*}{A G}  & F1 0 &  & 0.30 & 0.42 & 0.40 & 0.38 & 0.43 & - & 0.53 & 0.35 & 0.48 & 0.411 & \multirow{2}{*}{0.515}  \\ 
& & F1 1 &  & 0.73 & 0.61 & 0.60 & 0.68 & 0.55 & 0.82 & 0.49 & 0.56 & 0.53 &  0.619 &  \\ 
\cline{2-15}

& \multirow{2}{*}{M G}  & F1 0 &  & 0.37 & 0.43 & 0.43 & 0.40 & 0.46 & - & 0.71 & - & 0.46 &  0.466 & \multirow{2}{*}{\textbf{0.532}} \\ 
& & F1 1 &  & 0.71 & 0.64 & 0.61 & 0.70 & 0.55 & 0.78 & 0.11 & 0.78 & 0.51 & 0.599 & \\ 
\cline{2-15}
& \multirow{2}{*}{C FT} & F1 0 &  & 0.37 &  & 0.50 & 0.49 & 0.54 & - & 0.68 & 0.51 & 0.57 & 0.523 & \multirow{2}{*}{0.509} \\ 
& & F1 1 &  & 0.59 &  & 0.50 & 0.59 & 0.50 & 0.68 & 0.24 & 0.43 & 0.43 & 0.495 &  \\ 
\cline{2-15}

& \multirow{2}{*}{A G}  & F1 0 &  & 0.40 & 0.51 & 0.52 & 0.45 & 0.54 &  & 0.60 & 0.56 & 0.59 & 0.521 & \multirow{2}{*}{0.512} \\ 
& & F1 1 &  & 0.61 & 0.57 & 0.52 & 0.61 & 0.45 &  & 0.42 & 0.42 & 0.43 & 0.504 &  \\ 
\cline{2-15}

& \multirow{2}{*}{A G}  & F1 0 &  & 0.36 & 0.54 &  & 0.50 & 0.51 & - & 0.59 & 0.59 & 0.55 & 0.520 & \multirow{2}{*}{\textbf{0.535}} \\ 
& & F1 1 &  & 0.67 & 0.63 &  & 0.49 & 0.51 & 0.74 & 0.47 & 0.41 & 0.49 & 0.551 & \\ 
\cline{2-15}
& \multirow{2}{*}{A FT} & F1 0 &  & 0.40 & 0.43 & 0.43 & 0.37 &  & - & 0.55 & 0.26 & 0.50 & 0.420 & \multirow{2}{*}{0.522} \\ 
& & F1 1 &  & 0.72 & 0.64 & 0.58 & 0.72 &  & 0.77 & 0.47 & 0.59 & 0.51 & 0.625 & \\ 
\cline{2-15}
& \multirow{2}{*}{A G}  & F1 0 &  & 0.43 & 0.54 & 0.49 & 0.46 &  & - & 0.59 & 0.63 & 0.63 & 0.539 & \multirow{2}{*}{\textbf{0.539}} \\ 
& & F1 1 &  & 0.68 & 0.55 & 0.57 & 0.56 &  & 0.65 & 0.46 & 0.38 & 0.47 & 0.540 & \\ 
\hline
\end{tabular}
\arrayrulecolor{black}
\caption{Results on the datasets with attack (A) and support (S) relations. $F_1$ A stands for the $F_1$ measure of the attack relation and $F_1$ S stands for the $F_1$ measure of the support (S) relation. A stands for autoencoder model, C for concatenation model, M for mix model, G for GloVE embeddings, and FT for FastText embeddings. The blanks represent the training datasets. The Average (Avg) \OC{and the Macro (\OC{Mcr}) Avg do} not include the results of the dataset(s) used for training.}
\label{syntactic-100}
\end{table*}

\TC{Amongst the proposed architectures, the attention model generally performs better. Using GloVe embeddings instead of FastText yields better results. The autoencoder does not give good results, which may be attributed to the fact that using the encoding of the most salient features is not enough in predicting the argumentative relation and that analysing the entire sequence is better.}
\TC{Using only a single dataset for training, the models that perform the best are the \emph{attention} model and the \emph{mix} model, in both cases using \emph{all} features and trained on the \emph{essay} dataset.} 
\TC{The best results are obtained when using another dataset along one of the larger datasets for training. This is because combining data from two domains we are able to learn better the types of argumentative relations.} \TC{When using \emph{syntactic} features, adding \emph{micro}, \emph{cdcp}, and \emph{ukp} does not improve the results compared to using a single dataset for training. Indeed,  \emph{cdcp} has only one type of relation (i.e. support) resulting in an imbalanced dataset and in \emph{ukp}, the parent argument is a topic, which does not improve the prediction task. When using \emph{all} features, \emph{micro}, \emph{com}, \emph{ukp}, and \emph{nk} do not contribute to an increase in performance.}
\TC{The best results are obtained using the attention mechanism with GloVE embeddings trained on the \emph{web} and \emph{essay} datasets using \emph{syntactic} features (0.5445 macro average $F_1$).}

\subsection{Neural baselines with contextualised word embeddings}

\TC{Contextualised word embeddings such as the Bidirectional Encoder Representations from Transformers (BERT) embeddings \cite{bert} analyse the entire sentence before assigning an embedding to each word. The main difference between GloVE, FastText and contextualised word embeddings is that GloVE does not take the word order into account during training, whereas BERT do.}
\TC{We employ BERT embeddings to test whether they bring any improvements to the classification task.} \TC{While for GloVE/FastText vectors we do not need the \OC{original, trained} model in order to use the embeddings, for the contextualied word embeddings we require the pre-trained language models that we can then fine tune using the datasets of the downstream task.} \TC{We try different combinations for the neural network with BERT embeddings: using 3 or 4 BERT layers and using 1 dense layer (of 64 neurons) or 2 dense layers (of 128 and 32 neurons) before the final layer that determines the class.}
\TC{Table~\ref{bert} shows the results with BERT embeddings instead of Glove/FastText, following the same experiments described in Section \ref{non-context}: feature ablation (\emph{syntactic} vs \emph{all} features)  and using two datasets for training to test whether this can improve performance.}
\TC{The best results are obtained using 4 BERT layers and 2 dense layers (0.537 macro average $F_1$). However, \OC{the best BERT baseline} does not outperform the best results obtained using the attention model and GloVE.}

\begin{table*}
\centering
\arrayrulecolor{black}
\begin{tabular}{|l|l|l|c|c|c|c|c|c|c|c|c|c|c|c|}
\hline & & & essay & micro & db & ibm & com & web & cdcp & ukp & nk & aif & Avg & Mcr Avg\\ 
\hline

\multirow{22}{*}{\rotatebox[origin=c]{90}{BERT embeddings + syntactic}} & 3B & $F_1$ A &  & & 0.53 & 0.48 & 0.52 & 0.49 & - & 0.56 & 0.46 & 0.43 & 0.496 & \multirow{2}{*}{0.522}  \\ 
& 1D & $F_1$ S & & & 0.63 & 0.56 & 0.58 & 0.50 & 0.70 & 0.48 & 0.51 & 0.42 & 0.548 &  \\
\cline{2-15}
& 4B & $F_1$ A & & & 0.55 &	0.47 & 0.53 & 0.50 & -	& 0.56 & 0.48 & 0.45 & 0.506 & \multirow{2}{*}{0.526} \\
& 2D & $F_1$ S & & & 0.61&0.57&0.59&0.49&0.69&0.47&0.48 & 0.46 & 0.545 & \\
\cline{2-15}
& 4B & $F_1$ A & & 0.36&0.48&0.40&0.45&0.42&-&0.53 & & 0.37&0.430 & \multirow{2}{*}{0.525} \\
& 1D & $F_1$ S & & 0.69&0.67&0.61&0.62&0.57&0.79&0.50 & & 0.50&	0.619 & \\
\cline{2-15}
& 3B & $F_1$ A & & 0.39&0.57&0.53&0.46&0.53&-&0.61&0.57 & & 0.523 & \multirow{2}{*}{0.520} \\
& 1D & $F_1$ S & & 0.59&0.57&0.44&0.54&0.49&0.61&0.36&0.53 & & 0.516 &  \\
\cline{2-15}
& 4B & $F_1$ A & & 0.37&0.54&0.52&0.43&0.51&-&0.58&0.56 & & 0.501 & \multirow{2}{*}{0.521} \\
& 2D & $F_1$ S & & 0.61&0.58&0.45&0.57&0.52&0.65&0.40&0.55 & & 0.541 &  \\
\cline{2-15}
& 3B & $F_1$ A &  & 0.30&0.53& & 0.51&0.39&-&0.44&0.44&0.47&0.440 & \multirow{2}{*}{0.525} \\
& 2D & $F_1$ S & & 0.72&0.64& & 0.61&0.57&0.80&0.52&0.54&0.47&0.609 &  \\
\cline{2-15}
& 4B & $F_1$ A & & 0.33&0.49 & &  0.49&0.40&-&0.56&0.47&0.44&0.454 & \multirow{2}{*}{\textbf{0.531}} \\
& 1D & $F_1$ S & & 0.68&0.66 & & 0.61&0.56&0.78&0.46&0.56&0.55&0.608 &   \\
\cline{2-15}
& 4B & $F_1$ A &  & 0.29&0.49 & & 0.50&0.33&-&0.43&0.40&0.36&0.400 & \multirow{2}{*}{0.522} \\
& 2D & $F_1$ S &  & 0.72&0.68 & & 0.64&0.59&0.83&0.53&0.57&0.59&0.644 &  \\
\cline{2-15}
& 3B & $F_1$ A & 0.49 & 0.35&0.47& & 0.53&0.46& & 0.63&0.54&0.62&0.511 & \multirow{2}{*}{0.521} \\
& 2D & $F_1$ S & 0.53 & 0.63&0.55& & 0.64&0.50 & & 0.35&0.53&0.51&0.530 &  \\
\cline{2-15}
& 4B & $F_1$ A & 0.50 & 0.36&0.46& & 0.50 & & -&0.52&0.47&0.50&0.473 &\multirow{2}{*}{\textbf{0.537}} \\
& 2D & $F_1$ S & 0.61 & 0.62&0.59 & & 0.61 & & 0.74&0.52&0.50&0.61&0.600 &   \\
\cline{2-15}
& 4B & $F_1$ A & & 0.39&0.54&0.47&0.52 & & -&0.58&0.50&0.58&0.511 & \multirow{2}{*}{0.520} \\
& 1D & $F_1$ S & & 0.61&0.55&0.52&0.59 & & 0.69&0.46&0.48&0.32&0.528 &  \\
\hline

\multirow{8}{*}{\rotatebox[origin=c]{90}{all features}} & 3B & $F_1$ A & & 0.33&0.42&0.41&0.52&0.42&-&0.53&0.45&0.35&0.429 &  \multirow{2}{*}{0.524} \\
& 1D & $F_1$ S & & 0.67&0.66&0.63&0.63&0.58&0.80&0.52&0.54&0.53&0.618 &  \\
\cline{2-15} 
& 3B & $F_1$ A & & & 0.53&0.49&0.51&0.49&-&0.58&0.48&0.45&0.504 & \multirow{2}{*}{0.522} \\
& 1D & $F_1$ S & & & 0.62&0.56&0.61&0.50&0.69&0.46&0.48&0.40&0.540 &  \\
\cline{2-15} 
& 4B & $F_1$ A & & & 0.53&0.50&0.54&0.51&-&0.59&0.51&0.49&0.524 & \multirow{2}{*}{0.529} \\
& 1D & $F_1$ S & & & 0.59&0.56&0.55&0.47&0.67&0.45&0.48&0.49&0.533 &  \\
\cline{2-15}
& 3B & $F_1$ A & 0.48 & 0.34&0.48 & & 0.45 & &  -&0.45&0.50&0.54&0.463 & \multirow{2}{*}{\textbf{0.532}} \\
& 2D & $F_1$ S & 0.57 & 0.65&0.60 & & 0.64 & & 0.73&0.55&0.52&0.54&0.600 &   \\
\hline
\end{tabular}
\arrayrulecolor{black}
\caption{\TC{Results on the datasets with attack (A) and support (S) relations. $F_1$ A stands for the $F_1$ of the attack relation and $F_1$ S stands for the $F_1$ of the support (S) relation. \emph{X}B stands for the number of BERT layers used (i.e. \emph{X}) and \emph{Y}B stands for the number of dense layers (i.e. \emph{Y}) used before the final layer that predicts the class. The blanks represent the training datasets. The Average (Avg) \OC{and the Macro (Mcr) Avg do} not include the results of the dataset(s) used for training.}}
\label{bert}
\end{table*}

\subsection{Discussion}
Our baselines perform homogeneously over all existing datasets for relation prediction in AM while using generic features. 
\TC{As it may be noticed in the examples provided in Section~\ref{am-datasets}, the datasets differ at granularity: some consist of pairs of sentences (e.g., IBM) whereas others include pair of multiple-sentence arguments (e.g., Nixon-Kennedy debate). 
Additionally, the argumentation relations can be domain-specific and the semantic nature of argumentative relations may vary between corpora (e.g., ComArg). Thus, in this paper we considered a simpler but still complex task of determining the relation of either support or attack \OC{between two texts}.}
\TC{Embeddings represent the main difference in the features used for the machine learning models we experimented with.
Whilst word embeddings are often used as the first data processing layer
in a deep learning model, we employed TF-IDF features for the standard machine learning models that we considered as baselines.}
Other works that address the task of relation prediction make use of features specific to the single dataset of interest, making it difficult to test those models on the other datasets. For instance, for the \emph{essay} dataset, \newcite{Stab:17} use structural features such as number of preceding and following tokens in the covering sentence, number of components in paragraph, number of preceding and following components in paragraph, relative position of the argument component in paragraph. For the other datasets, \cite{Stab:18} use topic similarity features (as the \emph{parent} argument is a topic),  \cite{Menini:18} use the position of the topic and similarity with other related/unrelated pair from the dataset, keyword embeddings of topics from the dataset. \OC{We have used only general purpose features that are meaningful for all datasets addressing the relational AM task.}
\TC{Surprisingly, BERT embeddings that have achieved state-of-the-art in several tasks \cite{bert} do not bring any improvements compared to non-contextualised word embeddings for the relation prediction task in AM.}

\section{Conclusion} \label{concl}
Several resources have been built in the latest years for the task of argumentative relation prediction, covering different topics like political speeches, Wikipedia articles, persuasive essays. Given the heterogeneity of these kinds of text, it is hard to compare cross-dataset the different approaches proposed in the literature to address the argumentative relation prediction task. 
\TC{For this reason, in this paper, we addressed the issue of AM models that are hardly portable from one application \OC{dataset} to another due to the features used. We provided a broad comparison of different deep learning methods using both non-contextualised and contextualised word embeddings for a large set of datasets for the argumentative relation prediction, an important and still widely open problem.}  We proposed a set of strong dataset independent baselines based on several neural architectures and have shown that our models perform homogeneously over all existing datasets for relation prediction in AM.

\section{Bibliographical References}\label{reference}

\bibliographystyle{lrec}
\bibliography{lrec2020W-xample-kc}

\begin{thebibliography}{}

\bibitem[\protect\citename{Bar-Haim \bgroup et al.\egroup }2017]{Slonim:17}
Bar-Haim, R., Bhattacharya, I., Dinuzzo, F., Saha, A., and Slonim, N.
\newblock (2017).
\newblock Stance classification of context-dependent claims.
\newblock In {\em Proceedings of the 15th Conference of the European Chapter of
  the Association for Computational Linguistics: Volume 1, Long Papers}, pages
  251--261.

\bibitem[\protect\citename{Bex \bgroup et al.\egroup }2013]{Bex:13}
Bex, F., Modgil, S., Prakken, H., and Reed, C.
\newblock (2013).
\newblock On logical specifications of the argument interchange format.
\newblock {\em Journal of Logic and Computation}, 23(5):951--989.

\bibitem[\protect\citename{Boltu\v{z}i\'{c} and \v{S}najder}2014]{Boltuvzic:14}
Boltu\v{z}i\'{c}, F. and \v{S}najder, J.
\newblock (2014).
\newblock Back up your stance: Recognizing arguments in online discussions.
\newblock In {\em Proceedings of the First Workshop on Argumentation Mining},
  pages 49--58.

\bibitem[\protect\citename{Breiman}2001]{Breiman:01}
Breiman, L.
\newblock (2001).
\newblock Random forests.
\newblock {\em Machine Learning}, 45(1):5--32.

\bibitem[\protect\citename{Cabrio and Villata}2014]{Cabrio:14}
Cabrio, E. and Villata, S.
\newblock (2014).
\newblock Node: {A} benchmark of natural language arguments.
\newblock In {\em Computational Models of Argument - Proceedings of {COMMA}
  2014, Atholl Palace Hotel, Scottish Highlands, UK, September 9-12, 2014},
  pages 449--450.

\bibitem[\protect\citename{Cabrio and Villata}2018]{DBLP:conf/ijcai/CabrioV18}
Cabrio, E. and Villata, S.
\newblock (2018).
\newblock Five years of argument mining: a data-driven analysis.
\newblock In {\em Proceedings of the Twenty-Seventh International Joint
  Conference on Artificial Intelligence, {IJCAI}}, pages 5427--5433.

\bibitem[\protect\citename{Carstens and Toni}2015]{Carstens:15}
Carstens, L. and Toni, F.
\newblock (2015).
\newblock Towards relation based argumentation mining.
\newblock In {\em Proceedings of the 2nd Workshop on Argumentation Mining},
  pages 29--34.

\bibitem[\protect\citename{Carstens and Toni}2017]{Carstens:17}
Carstens, L. and Toni, F.
\newblock (2017).
\newblock Using argumentation to improve classification in natural language
  problems.
\newblock {\em {ACM} Transactions on Internet Technology}, 17(3):30:1--30:23.

\bibitem[\protect\citename{Chang and Lin}2011]{Chang:11}
Chang, C. and Lin, C.
\newblock (2011).
\newblock {LIBSVM:} {A} library for support vector machines.
\newblock {\em {ACM} Transactions on Intelligent Systems and Technology
  {TIST}}, 2(3):27:1--27:27.

\bibitem[\protect\citename{Ches{\~{n}}evar \bgroup et al.\egroup
  }2006]{Chesnevar:06}
Ches{\~{n}}evar, C.~I., McGinnis, J., Modgil, S., Rahwan, I., Reed, C., Simari,
  G.~R., South, M., Vreeswijk, G., and Willmott, S.
\newblock (2006).
\newblock Towards an argument interchange format.
\newblock {\em Knowledge Engineering Review}, 21(4):293--316.

\bibitem[\protect\citename{Cho \bgroup et al.\egroup }2014]{Cho:14}
Cho, K., van Merri{\"{e}}nboer, B., G{\"{u}}l{\c c}ehre, {\c C}., Bahdanau, D.,
  Bougares, F., Schwenk, H., and Bengio, Y.
\newblock (2014).
\newblock Learning phrase representations using {RNN} {E}ncoder--{D}ecoder for
  statistical machine translation.
\newblock In {\em Proceedings of the 2014 Conference on Empirical Methods in
  Natural Language Processing (EMNLP)}, pages 1724--1734.

\bibitem[\protect\citename{Chung \bgroup et al.\egroup }2014]{Chung:14}
Chung, J., G{\"{u}}l{\c{c}}ehre, {\c{C}}., Cho, K., and Bengio, Y.
\newblock (2014).
\newblock Empirical evaluation of gated recurrent neural networks on sequence
  modeling.
\newblock {\em CoRR}, abs/1412.3555.

\bibitem[\protect\citename{Devlin \bgroup et al.\egroup }2018]{bert}
Devlin, J., Chang, M., Lee, K., and Toutanova, K.
\newblock (2018).
\newblock {BERT:} pre-training of deep bidirectional transformers for language
  understanding.
\newblock {\em CoRR}, abs/1810.04805.

\bibitem[\protect\citename{Erhan \bgroup et al.\egroup }2010]{Erhan:10}
Erhan, D., Bengio, Y., Courville, A., Manzagol, P.-A., Vincent, P., and Bengio,
  S.
\newblock (2010).
\newblock Why does unsupervised pre-training help deep learning?
\newblock {\em Journal of Machine Learning Research}, 11:625--660.

\bibitem[\protect\citename{Esuli and Sebastiani}2006]{Esuli:06}
Esuli, A. and Sebastiani, F.
\newblock (2006).
\newblock Sentiwordnet: A publicly available lexical resource for opinion
  mining.
\newblock In {\em In Proceedings of the 5th Conference on Language Resources
  and Evaluation (LREC'06}, pages 417--422.

\bibitem[\protect\citename{Habernal and
  Gurevych}2017]{Habernal:2017:AMU:3097255.3097259}
Habernal, I. and Gurevych, I.
\newblock (2017).
\newblock Argumentation mining in user-generated web discourse.
\newblock {\em Computational Linguistics}, 43(1):125--179.

\bibitem[\protect\citename{Han \bgroup et al.\egroup }2017]{Han:17}
Han, K., Li, C., and Shi, X.
\newblock (2017).
\newblock Autoencoder feature selector.
\newblock {\em CoRR}, abs/1710.08310.

\bibitem[\protect\citename{Hinton and Salakhutdinov}2006]{Hinton:06}
Hinton, G. and Salakhutdinov, R.
\newblock (2006).
\newblock Reducing the dimensionality of data with neural networks.
\newblock {\em Science}, 313(5786):504 -- 507.

\bibitem[\protect\citename{Hutto and Gilbert}2014]{Hutto:14}
Hutto, C.~J. and Gilbert, E.
\newblock (2014).
\newblock {VADER:} {A} parsimonious rule-based model for sentiment analysis of
  social media text.
\newblock In {\em Proceedings of the Eighth International Conference on Weblogs
  and Social Media, {ICWSM}}.

\bibitem[\protect\citename{Iyad and Reed}2009]{Rahwan:09}
Iyad, R. and Reed, C.
\newblock (2009).
\newblock The argument interchange format.
\newblock In {\em Argumentation in Artificial Intelligence}, pages 383--402.

\bibitem[\protect\citename{Joulin \bgroup et al.\egroup }2016]{Joulin:16}
Joulin, A., Grave, E., Bojanowski, P., Douze, M., J{\'e}gou, H., and Mikolov,
  T.
\newblock (2016).
\newblock Fasttext.zip: Compressing text classification models.
\newblock {\em arXiv preprint arXiv:1612.03651}.

\bibitem[\protect\citename{J{\'{o}}zefowicz \bgroup et al.\egroup
  }2015]{Jozefowicz:15}
J{\'{o}}zefowicz, R., Zaremba, W., and Sutskever, I.
\newblock (2015).
\newblock An empirical exploration of recurrent network architectures.
\newblock In {\em Proceedings of the 32nd International Conference on Machine
  Learning, {ICML}}, pages 2342--2350.

\bibitem[\protect\citename{Lawrence \bgroup et al.\egroup }2012]{Lawrence:12}
Lawrence, J., Bex, F., Reed, C., and Snaith, M.
\newblock (2012).
\newblock Aifdb: Infrastructure for the argument web.
\newblock In {\em Computational Models of Argument - Proceedings of {COMMA}},
  volume 245, pages 515--516.

\bibitem[\protect\citename{Lippi and Torroni}2016]{DBLP:journals/toit/LippiT16}
Lippi, M. and Torroni, P.
\newblock (2016).
\newblock Argumentation mining: State of the art and emerging trends.
\newblock {\em {ACM} Transactions on Internet Technology}, 16(2):10.

\bibitem[\protect\citename{Menini \bgroup et al.\egroup }2018]{Menini:18}
Menini, S., Cabrio, E., Tonelli, S., and Villata, S.
\newblock (2018).
\newblock Never retreat, never retract: Argumentation analysis for political
  speeches.
\newblock In {\em Proceedings of the Thirty-Second {AAAI} Conference on
  Artificial Intelligence}, pages 4889--4896.

\bibitem[\protect\citename{Mikolov \bgroup et al.\egroup }2018]{Mikolov:18}
Mikolov, T., Grave, E., Bojanowski, P., Puhrsch, C., and Joulin, A.
\newblock (2018).
\newblock Advances in pre-training distributed word representations.
\newblock In {\em Proceedings of the International Conference on Language
  Resources and Evaluation (LREC)}.

\bibitem[\protect\citename{Miller}1995]{Miller:95}
Miller, G.~A.
\newblock (1995).
\newblock Wordnet: A lexical database for english.
\newblock {\em Communications of the ACM}, 38:39--41.

\bibitem[\protect\citename{Mochales and Moens}2011]{Mochales2011}
Mochales, R. and Moens, M.-F.
\newblock (2011).
\newblock Argumentation mining.
\newblock {\em Artificial Intelligence and Law}, 19(1):1--22.

\bibitem[\protect\citename{Niculae \bgroup et al.\egroup }2017]{Niculae:17}
Niculae, V., Park, J., and Cardie, C.
\newblock (2017).
\newblock Argument mining with structured {SVM}s and {RNN}s.
\newblock In {\em Proceedings of the 55th Annual Meeting of the Association for
  Computational Linguistics (Volume 1: Long Papers)}, pages 985--995.

\bibitem[\protect\citename{Parikh \bgroup et al.\egroup }2016]{Parikh:16}
Parikh, A.~P., T{\"a}ckstr{\"o}m, O., Das, D., and Uszkoreit, J.
\newblock (2016).
\newblock A decomposable attention model for natural language inference.
\newblock In {\em EMNLP}.

\bibitem[\protect\citename{Park and Cardie}2018]{Park:18}
Park, J. and Cardie, C.
\newblock (2018).
\newblock A corpus of e{R}ulemaking user comments for measuring evaluability of
  arguments.
\newblock In {\em Proceedings of the Eleventh International Conference on
  Language Resources and Evaluation, {LREC}}.

\bibitem[\protect\citename{Peldszus and
  Stede}2013]{DBLP:journals/ijcini/PeldszusS13}
Peldszus, A. and Stede, M.
\newblock (2013).
\newblock From argument diagrams to argumentation mining in texts: {A} survey.
\newblock {\em {IJCINI}}, 7(1):1--31.

\bibitem[\protect\citename{Peldszus and Stede}2015]{Peldszus:15}
Peldszus, A. and Stede, M.
\newblock (2015).
\newblock Joint prediction in mst-style discourse parsing for argumentation
  mining.
\newblock In {\em Proceedings of the 2015 Conference on Empirical Methods in
  Natural Language Processing, {EMNLP}}, pages 938--948.

\bibitem[\protect\citename{Pennington \bgroup et al.\egroup
  }2014]{Pennington:14}
Pennington, J., Socher, R., and Manning, C.~D.
\newblock (2014).
\newblock Glove: Global vectors for word representation.
\newblock In {\em Empirical Methods in Natural Language Processing (EMNLP)},
  pages 1532--1543.

\bibitem[\protect\citename{Reed \bgroup et al.\egroup }2008]{Reed:08a}
Reed, C., Wells, S., Devereux, J., and Rowe, G.
\newblock (2008).
\newblock {AIF+:} dialogue in the argument interchange format.
\newblock In {\em Computational Models of Argument: Proceedings of {COMMA}},
  pages 311--323.

\bibitem[\protect\citename{Stab and Gurevych}2017]{Stab:17}
Stab, C. and Gurevych, I.
\newblock (2017).
\newblock Parsing argumentation structures in persuasive essays.
\newblock {\em Computational Linguistics}, 43(3):619--659.

\bibitem[\protect\citename{Stab \bgroup et al.\egroup }2018]{Stab:18}
Stab, C., Miller, T., Schiller, B., Rai, P., and Gurevych, I.
\newblock (2018).
\newblock Cross-topic argument mining from heterogeneous sources.
\newblock In {\em Proceedings of the Conference on Empirical Methods in Natural
  Language Processing (EMNLP)}.

\bibitem[\protect\citename{Teruel \bgroup et al.\egroup }2018]{Teruel:18}
Teruel, M., Cardellino, C., Cardellino, F., Alemany, L.~A., and Villata, S.
\newblock (2018).
\newblock Increasing argument annotation reproducibility by using
  inter-annotator agreement to improve guidelines.
\newblock In {\em Proceedings of the Eleventh International Conference on
  Language Resources and Evaluation, {LREC}}.

\bibitem[\protect\citename{Vaswani \bgroup et al.\egroup }2017]{Vaswani:17}
Vaswani, A., Shazeer, N., Parmar, N., Uszkoreit, J., Jones, L., Gomez, A.~N.,
  Kaiser, L., and Polosukhin, I.
\newblock (2017).
\newblock Attention is all you need.
\newblock In {\em Advances in Neural Information Processing Systems 30: Annual
  Conference on Neural Information Processing Systems}, pages 6000--6010.

\bibitem[\protect\citename{Walker \bgroup et al.\egroup }2012]{Walker:12}
Walker, M.~A., Tree, J. E.~F., Anand, P., Abbott, R., and King, J.
\newblock (2012).
\newblock A corpus for research on deliberation and debate.
\newblock In {\em Proceedings of the Eighth International Conference on
  Language Resources and Evaluation, {LREC}}, pages 812--817.

\bibitem[\protect\citename{Wang \bgroup et al.\egroup }2017]{Wang:17}
Wang, S., Ding, Z., and Fu, Y.
\newblock (2017).
\newblock Feature selection guided auto-encoder.
\newblock In {\em Proceedings of the Thirty-First {AAAI} Conference on
  Artificial Intelligence}, pages 2725--2731.

\bibitem[\protect\citename{Yang \bgroup et al.\egroup }2016]{Yang:16}
Yang, Z., Yang, D., Dyer, C., He, X., Smola, A.~J., and Hovy, E.~H.
\newblock (2016).
\newblock Hierarchical attention networks for document classification.
\newblock In {\em {NAACL} {HLT} 2016, The 2016 Conference of the North American
  Chapter of the Association for Computational Linguistics: Human Language
  Technologies}, pages 1480--1489.

\end{thebibliography}

\end{document}